\documentclass{article}
\newif\ifpaperlocal
\IfFileExists{style/iclr2027_conference.sty}{\paperlocaltrue}{\paperlocalfalse}
\makeatletter
\def\input@path{{paper/}}
\makeatother
\ifpaperlocal
  \usepackage{style/iclr2027_conference}
\else
  \usepackage{paper/style/iclr2027_conference}
\fi
\usepackage{times}
\iclrfinalcopy
\setcitestyle{numbers,square}
\usepackage[utf8]{inputenc}
\usepackage[T1]{fontenc}
\usepackage{hyperref}
\usepackage{url}
\usepackage{booktabs}
\usepackage{amsmath,amssymb}
\usepackage{graphicx}
\graphicspath{{figures/}{../figures/}}
\usepackage{microtype}
\usepackage{xcolor}
\usepackage{array}
\usepackage{longtable}
\usepackage{placeins}
\newcolumntype{L}[1]{>{\raggedright\arraybackslash}p{#1}}
\hypersetup{hidelinks,pdftitle={Lean Pool: An AI-Maintained Archive of Formalized Mathematics},pdfauthor={Vasily Ilin}}
\newcommand{\pool}{Lean Pool}
\newcommand{\code}[1]{\texttt{#1}}
\title{Lean Pool: An AI-Maintained Archive of\\Formalized Mathematics}
\author{Vasily Ilin\\University of Washington}

\begin{document}
\maketitle
\lhead{Preprint}
\begin{abstract}
Lean Pool is a repository of formalized mathematics. It is grown, maintained and
optimized by AI agents.
\end{abstract}

\begin{quote}
\small\noindent\textbf{Disclaimer.}
The human-written portion of this paper consists of a single page. The author
believes that it's enough to convey the main idea. The rest of the paper is
produced almost entirely by AI.
\end{quote}

\section{Lean Pool}
Generative AI accounts for most of the recent AI progress, including the incredible recent advancements in mathematics. As generation becomes commodified, verification becomes the bottleneck. Lean~\citep{moura2021lean4} has emerged as the primary language to verify mathematical proof, both human-made and AI-generated. However, Lean's standard math library, Mathlib~\citep{mathlib2020,baanen2026growing}, lacks the definitions and theorems needed to formalize much of research-level mathematics. Concerningly, Mathlib continues to grow at a linear rate due to the strict human review. 

We introduce Lean Pool, a repository of formalized mathematics, which is grown, maintained, and optimized by AI agents. The Lean kernel guarantees correctness of proofs, and a combination of strict linters and LLM review aim to uphold the quality of definitions and theorem statements. Additionally, Lean Pool's codebase is regularly optimized for conciseness, compilation speed and RAM usage.

\paragraph{Growth.} Lean Pool is grown in two primary ways. First, by AI agents discovering formalization projects under Apache-2.0 or MIT licenses and pooling them. Second, by human contributors pooling their projects.
Only serious complete formalizations of named known results are eligible to be pooled. Both human-written and AI-generated projects are eligible. Pooling involves bumping the Lean version, making the Pull Request pass the Continuous Integration linters and the LLM reviewer, and optimizing the hotspots for lower compilation time and RAM.

\paragraph{Maintenance.} Lean Pool is maintained in two ways. First, when Mathlib's version changes, an AI agent bumps Lean Pool's version and resolves the errors. Second, AI agents regularly optimize the codebase for conciseness, compilation speed and RAM usage.

\paragraph{Documentation.} Lean Pool has three types of documentation: the traditional Index\footnote{index: https://vilin97.github.io/lean-pool/}, Exposition\footnote{Exposition: https://vilin97.github.io/lean-pool/exposition/}, and daily project announcements in Zulip\footnote{Announcements: \url{https://leanprover.zulipchat.com/\#narrow/channel/619231-Lean-Pool}}.

\paragraph{Statistics.} At the time of writing, Lean Pool contains 211 pooled projects. They comprise 3,228,485 lines of Lean code. There are 18 contributors. The Lean version has been bumped six times.

\paragraph{Vision.} As formalization becomes easier and new math results are immediately formalized upon release, Lean Pool can serve as the formal analogue of the arXiv.org website -- a place to quickly share new work, with minimal friction.

\clearpage
\subsection*{Extended summary}
As AI systems increasingly contribute to mathematical research, formal proofs make
it possible to check arguments before they have been fully examined by
mathematicians. Reusing these proofs requires more than preserving their source:
formalizations must remain compatible with evolving libraries, and their results
must be easy to find and understand. We present Lean Pool, a living archive of
mathematical formalizations maintained together as Lean and Mathlib evolve. It
combines AI agents, automated checks, and human oversight to maintain independently
developed projects while preserving their attribution. We analyze the archive's
contribution and maintenance history, accepted optimizations, mathematical reviews,
and evidence of reuse. The operational record shows that agent-assisted maintenance
can restore compatibility across dependency upgrades and support library-wide proof
shortening and compilation improvements. It also records follow-up repairs after
initial automation, the resource demands of large reviews, and tradeoffs between
reusable interfaces and compilation cost. Archived mathematics is reused in
subsequent research-level formalization. We provide the archive, its maintenance
workflows, and exposition site, aiming to provide a
maintained formal counterpart to arXiv.

\section{Motivation and scope}
AI systems can produce mathematical arguments faster than mathematicians can read
and absorb them. Formalization makes the correctness of these arguments mechanically
checkable while their ideas are still being understood. OpenAI's \emph{Ten Advances
in Mathematics and Theoretical Computer Science} and \emph{Finite Time Blowup for
Navier--Stokes} illustrate this role: both released long AI-generated arguments
alongside Lean formalizations~\citep{poolpub_ten_advances,openai2026navier}.
The releases describe how formalization accompanied these discoveries~\citep{openai2026tenrelease,openai2026navierrelease}.

These formalizations can also become foundations for later research. A reader needs
to find the relevant theorem, inspect its assumptions, and use it in a new development.
That requires more than preserving the original source: as Lean and Mathlib evolve,
the formalization must remain compatible with the libraries on which new work depends.

\href{https://github.com/Vilin97/lean-pool}{\pool{}} is a living archive for this purpose. It gives completed formalizations a
persistent, searchable home while maintaining them together in a common Lean/Mathlib
environment. Its contributors include people submitting their own mathematics,
people using AI agents, and agents discovering and importing existing projects.
The archive preserves attribution and project organization while applying shared
checks to contributions and subsequent maintenance.

\paragraph{Admission rules.}
Completed projects must contain no \code{sorry} or \code{admit} and introduce no axioms
beyond \code{Classical.choice}, \code{propext}, and \code{Quot.sound}.
They must avoid \code{set\_option}, unchecked declarations, and mechanisms that bypass
the repository's resource limits or linters. Every project requires a card identifying
its authors, upstream source, proof provenance, and main results with their informal
statements. Accepted source must carry an \code{Apache-2.0} or \code{MIT} license.
The mechanical checks appear in
Appendix~\ref{app:ci}. Open challenges are kept separately from completed formalizations.

\begin{figure}[htbp]
\centering\includegraphics[width=\linewidth]{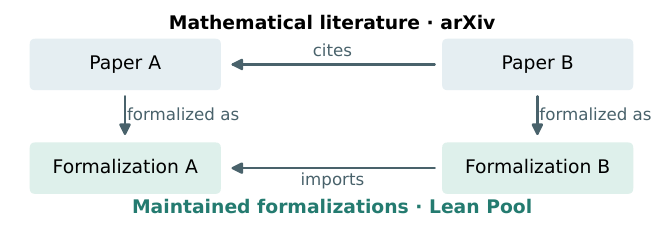}
\caption{The proposed relationship between arXiv and Lean Pool. New mathematics has
both a paper and a maintained formalization. An author places the paper on arXiv and
contributes its formalization to Lean Pool. Subsequent work cites the paper and imports
the formalization, so formal dependencies can mirror the dependency graph of the
mathematical literature. As formalization becomes easier and cheaper, we anticipate
that most new mathematics papers will have accompanying formalizations. Lean Pool
provides a home for these developments and maintains their connections as Lean and
Mathlib evolve.}
\label{fig:mirror}
\end{figure}

\paragraph{A formal counterpart to the mathematical literature.}
Figure~\ref{fig:mirror} illustrates our proposal for connecting mathematical papers
to a growing collection of maintained formalizations.

This paper analyzes the archive and the operations already used to maintain it:
community contributions, dependency upgrades, accepted optimization changes, deployed
LLM reviews, and the tools through which readers find and inspect mathematics.
Historical builds show that agent-assisted upgrades restore compatibility.
Accepted changes demonstrate library-wide proof shortening and faster compilation.
A separate LeanEval audit documents reuse of archived mathematics in research-level
formalization~\citep{leanevalanalysis2026}. We provide the archive, its maintained
workflows, and its exposition site.

\section{Contributing to and maintaining the archive}\label{sec:system}
\paragraph{Community contributions.}
Work enters \pool{} through direct contributions and through attributed imports of
upstream projects. A contributor can propose a repository or submit a prepared project.
The PR history includes contributions of new formalizations, improvements to archived
proofs, and infrastructure for discovering and checking projects.
Appendix~\ref{app:community} attributes these contributions.

A submitter and a formalization's authors need not be the same people. Project cards
retain upstream authorship even when a maintainer or an agent performs the import.
Their provenance labels distinguish human-written, AI-written, and mixed proofs.
GitHub accounts identify who submitted a change; they do not measure the division of
labor between that person and their agents.

\paragraph{Recurring work.}
Daily jobs search for recent and older formalizations, inspect open PRs, address
maintainer issues, optimize existing projects, and announce accepted contributions.
A separate dependency-update workflow detects new Lean/Mathlib releases, builds the
archive, assigns failing projects to repair agents, and assembles their patches for
review. These job definitions evolve alongside the archive.
Appendix~\ref{app:daily-jobs} describes their responsibilities and outputs.

\paragraph{Continuous integration.}
CI combines a full-library build, warning checks,
Mathlib's declaration and source-style linters, and archive-specific quality gates.
The latter check project cards, attribution, allowed axioms, proof and file sizes,
and attempts to bypass the checks. A compiled-environment audit complements source
scanning. Profiling reports the compilation cost of new files and compares modified
files with their earlier versions; it is advisory in the observed workflow.

These checks share Mathlib's build-and-lint foundation. \pool{} adds admission rules
for independently attributed projects and LLM review of their mathematical claims.
Tau Ceti also uses Mathlib linters and an axiom audit, and its performance workflow
makes resource regression checks a merge condition. Appendix~\ref{app:ci} compares the
mechanical checks and profiling methods directly.

\paragraph{Evidence and scope.}\label{sec:methods}
We analyze dated observations of a continuously changing archive, together with its
source history and retained public PR records. Compatibility evidence combines upgrade replays with the production upgrade
logs. Optimization results describe accepted changes, and review
statistics describe retained service reports. The build-resource comparison
uses fresh clean library builds on the same machine. Appendix~\ref{app:methods} specifies the observation
periods and measurement scopes.

\section{The archive and evidence of reuse}\label{sec:corpus}
The collection spans logic, number theory, algebra, analysis, geometry, probability,
and computer science, with scale and participation summarized in Table~\ref{tab:archive-profile}.
\begin{table}[htbp]
\centering\small\setlength{\tabcolsep}{4pt}\renewcommand{\arraystretch}{1.08}
\caption{Scale and participation at the source observation specified in Appendix~\ref{app:methods}. Physical source lines include comments and blank lines. Declaration commands are counted in source, excluding examples and generated auxiliaries. Project-card provenance describes proof authorship. Contributor counts use GitHub user accounts and exclude bots; commit contributors and merged-PR authors are counted separately. Community PRs exclude the maintainer account. Open challenges are outside the source census.}
\label{tab:archive-profile}
\begin{tabular}{lr}
\toprule
Archive property & Count\\
\midrule
Completed projects & 211\\
Lean source files & 7,043\\
Physical source lines & 3,228,485\\
Source declaration commands & 193,862\\
Registered main results & 837\\
Human / AI / mixed projects & 70 / 102 / 39\\
Commit contributors & 18\\
Contributors with merged PRs & 17\\
Community PRs merged & 63\\
\bottomrule
\end{tabular}
\end{table}

Its contents range from classical structural theorems to recently proved results.
The classification of compact surfaces development builds from triangulations to
normal forms for surfaces with boundary. The incompleteness development formalizes
G\"odel's theorems for arithmetic, together with arithmetization and provability logic.
The polynomial Freiman--Ruzsa development uses entropy to establish bounds in additive
combinatorics and reuses the archive's existing entropy library.
The Kurosh subgroup development likewise builds on archived results about
fundamental groups of finite graphs.
Other developments include the Navier--Stokes and Euler blowup developments,
non-sofic groups, and results on quantum parallel repetition
\citep{openai2026navier,openai2026euler,poolpub_ten_advances}.
The Komlós development formalizes the vector-balancing bound and its Beck--Fiala
discrepancy consequence~\citep{karingula2026komlos}.
The archive also contains a characterization of language generation in the limit,
connecting formalized mathematics to learning theory
\citep{poolupdate_language_generation_characterization}.
Appendix~\ref{app:project-papers} links the imported projects to their publications;
the \href{https://leanprover.zulipchat.com/\#narrow/channel/619231-Lean-Pool}{Lean Pool
Zulip channel} announces newly added results with attribution.

\begin{figure}[htbp]
\centering\includegraphics[width=\linewidth]{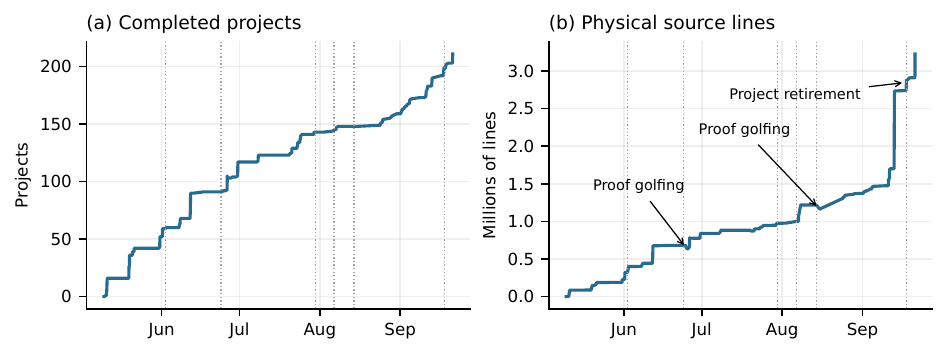}
\caption{Growth of the living archive. The curves show registered projects and
physical Lean source lines, including comments and blank lines. Dotted markers
indicate dependency upgrades. The project count continues to grow through periods
of proof golfing; the annotated dips in source size mark library-wide compression
and compilation optimizations. A separate project removal is labeled explicitly.}
\label{fig:growth}
\end{figure}

Figure~\ref{fig:growth} follows the archive's growth and maintenance over time.

\paragraph{Reuse in research-level mathematics.}
Lean Pool is the most reused external repository in the LeanEval structural
audit~\citep{leanevalanalysis2026}; Figure~\ref{fig:leaneval-reuse} in the appendix
reports the comparison and its definition of reuse.

\paragraph{Build resources.}
Appendix~\ref{app:build-methods} compares clean builds of the current archive,
Mathlib, and Tau Ceti on the same Azure host, with dependency caches prepared before timing.

\FloatBarrier
\section{Maintaining compatibility as dependencies evolve}\label{sec:maintenance}
A common environment remains useful only if archived projects can move with Lean and
Mathlib. The dependency-update workflow tests unchanged source under new dependencies,
assigns failures to agents, and checks the repaired projects together.
Table~\ref{tab:maintenance-main} records the projects affected by historical upgrades.
\begin{table}[htbp]
\centering\small\setlength{\tabcolsep}{4pt}\renewcommand{\arraystretch}{1.08}
\caption{Projects encountering compiler failures under dependency upgrades. Earlier rows replay the original pre-upgrade source with the new environment. The final row uses the retained production probe before repairs. It measures the archive at that probe; subsequent imports and a separately accepted curation change altered membership before merge. The stable migration ultimately passed the combined archive build and repository checks. Warning-only cleanup is excluded from the failure count.}
\label{tab:maintenance-main}
\begin{tabular}{llrr}
\toprule
From Lean & To Lean & Projects at probe & Compiler failures\\
\midrule
4.30.0-rc2 & 4.31.0-rc1 & 59 & 44\\
4.31.0-rc1 & 4.32.0-rc1 & 91 & 58\\
4.32.0-rc1 & 4.33.0-rc1 & 143 & 100\\
4.33.0-rc1 & 4.33.0-rc2 & 145 & 19\\
4.33.0-rc2 & 4.34.0-rc1 & 148 & 3\\
4.34.0-rc1 & 4.34.0 & 191 & 97\\
\bottomrule
\end{tabular}
\end{table}

The original upstream environments, including earlier Lean releases, are listed for
every imported project in Appendix~\ref{app:source-versions}.
The migration to stable Lean required follow-up integration after the initial
repair jobs, including updates to supporting APIs; Appendix~\ref{app:migration}
records the production stages and source changes as a proxy for maintenance effort.

\section{Proof shortening and compilation speed}\label{sec:optimization}
A shared archive also permits improvements across independently developed projects.
Accepted changes include library-wide proof compression, contributor-supplied golfing,
replacement of expensive proof searches, and simplification of computation-heavy
certificates. Table~\ref{tab:optimization-main} measures their effect on the source size,
complete-library build time, and peak memory at their historical revisions.
\begin{table}[!t]
\centering\small\setlength{\tabcolsep}{4pt}\renewcommand{\arraystretch}{1.10}
\caption{Clean builds of the complete Lean Pool library before and after accepted PRs. Dependencies and build settings are fixed within each pair on the same Azure VM. Time is elapsed build time; RAM is sampled peak combined memory of the build workers. Parentheses give time saved, computed from unrounded measurements; negative values mean longer builds. Source savings include refactoring and declaration removal. Proof shortening does not uniformly reduce build time: contributor golfing uses less memory but takes longer in this run. Each row reports one clean before/after pair; Appendix~\ref{app:optimization} gives the protocol.}
\label{tab:optimization-main}
\begin{tabular}{L{.31\linewidth}rcc}
\toprule
Accepted change & \shortstack{Lines removed} & \shortstack{Build min $\downarrow$\\before $\to$ after} & \shortstack{RAM GiB $\downarrow$\\before $\to$ after}\\
\midrule
\href{https://github.com/Vilin97/lean-pool/pull/117}{117} Proof-search simplification & 511 & \shortstack{5.30 $\to$ 5.25\\(+1.1\%)} & 15.6 $\to$ 15.6\\
\href{https://github.com/Vilin97/lean-pool/pull/187}{187} Library-wide compression & 45,217 & \shortstack{16.11 $\to$ 15.59\\(+3.3\%)} & 26.1 $\to$ 26.9\\
\href{https://github.com/Vilin97/lean-pool/pull/246}{246} Contributor proof golfing & 12,515 & \shortstack{21.83 $\to$ 23.01\\(-5.4\%)} & 25.9 $\to$ 24.6\\
\href{https://github.com/Vilin97/lean-pool/pull/338}{338} Certificate simplification & 513 & \shortstack{30.22 $\to$ 29.13\\(+3.6\%)} & 21.0 $\to$ 20.9\\
\href{https://github.com/Vilin97/lean-pool/pull/339}{339} Elaboration-cost reduction & 54,965 & \shortstack{28.46 $\to$ 26.82\\(+5.8\%)} & 19.8 $\to$ 19.6\\
\bottomrule
\end{tabular}
\end{table}

Table~\ref{tab:recent-optimizations} reports import cleanup, proof simplification,
and reusable-API work, using the project-level measurements recorded with those changes.
\begin{table}[htbp]
\centering\small\setlength{\tabcolsep}{4pt}\renewcommand{\arraystretch}{1.08}
\caption{Accepted project-level optimizations and API work. The proof optimizations reduce project compilation time, while the Navier--Stokes refactor adds reusable interfaces at a small compilation cost. Positive source reductions mean fewer lines; the API import has no comparable source-reduction baseline. Timings come from the linked PRs, with dependencies prepared before each measured build. Workloads and machines differ across rows. Import cleanup has unmatched cache preparation, so only its source reduction is shown. Measurement summaries appear in Appendix~\ref{app:optimization}.}
\label{tab:recent-optimizations}
\begin{tabular}{lrr}
\toprule
Accepted change & Net lines removed & Project seconds $\downarrow$\\
\midrule
\href{https://github.com/Vilin97/lean-pool/pull/419}{Tactic-import cleanup} & 735 & ---\\
\href{https://github.com/Vilin97/lean-pool/pull/441}{Restricted-sum proofs} & 65 & 20.01 $\rightarrow$ 18.45\\
\href{https://github.com/Vilin97/lean-pool/pull/405}{Navier--Stokes reusable APIs} & --- & 879.67 $\rightarrow$ 887.78\\
\href{https://github.com/Vilin97/lean-pool/pull/447}{Infinite Connes rigidity} & -31 & 118.00 $\rightarrow$ 93.70\\
\href{https://github.com/Vilin97/lean-pool/pull/454}{Burkholder majorant proofs} & 1,064 & 23.84 $\rightarrow$ 19.35\\
\href{https://github.com/Vilin97/lean-pool/pull/459}{Quantum parallel repetition} & 32 & 141.85 $\rightarrow$ 86.08\\
\href{https://github.com/Vilin97/lean-pool/pull/465}{Interior-point LP proofs} & 249 & 13.98 $\rightarrow$ 10.69\\
\bottomrule
\end{tabular}
\end{table}

The archive has also adopted Lean modules and narrower imports alongside shared
proof arguments; Appendix~\ref{app:optimization} reports the accepted change's
historical fixed-workload benchmark.

\FloatBarrier
\section{Mathematical review in operation}\label{sec:review}
\begin{figure}[htbp]
\centering\includegraphics[width=\linewidth]{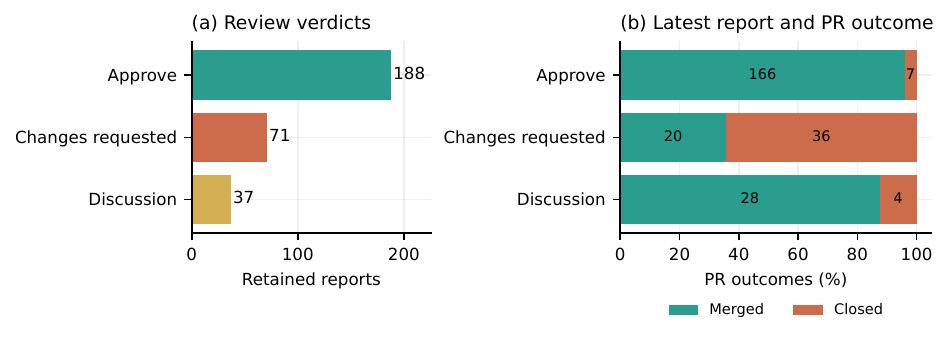}
\caption{Retained mathematical review-service reports. Left: all retained structured
reports. Right: PR outcomes at the observation date, grouped by each PR's latest
retained verdict; labels give counts and bar lengths give proportions. Most reports
approve their submissions. Requests for changes and discussion verdicts are each
followed by both closures and subsequent merges. All reviewed PRs have reached one
of these outcomes by the observation date. Findings concern
statement mismatches, attribution, incomplete results, duplicate definitions, and
unnecessary code. Greptile's advisory reviews and free-form daily-agent comments are
outside this census. The recorded outcomes are merge and closure decisions;
review accuracy was not independently labeled.}
\label{fig:production-reviews}
\end{figure}

A proof checker validates a formal statement, while admission also requires that the
statement corresponds to the contribution being advertised. The mathematical review
service has examined faithfulness, novelty, significance, sources, and code quality.
Maintainers and contributors can revise the submission or resolve questions raised
by its report. Figure~\ref{fig:production-reviews} summarizes the service's recorded
verdicts and subsequent PR outcomes.

Table~\ref{tab:review-costs} summarizes recorded price estimates; repeated-review agreement
and the service's subsequent redesign and pause are reported in Appendix~\ref{app:reviews}.
\begin{table}[htbp]
\centering\small\setlength{\tabcolsep}{4pt}\renewcommand{\arraystretch}{1.08}
\caption{Recorded review price estimates, separated by billing regime. Historical comments report API-based estimates. The newer Azure service consumes Codex account quota and reports a Standard-API-equivalent price using uncached-input rates. The equivalent prices are not cash payments. Totals cover retained reports with recorded estimates and exclude missing or overwritten executions.}
\label{tab:review-costs}
\begin{tabular}{lrrr}
\toprule
Price basis & Reports & Total & Median\\
\midrule
Historical API estimates & 285 & \$308.52 & \$0.20\\
Codex API-equivalent estimates & 6 & \$752.38 & \$89.03\\
\bottomrule
\end{tabular}
\end{table}

\FloatBarrier
\section{Exposition and the structure of the library}\label{sec:inspection}
The \href{https://vilin97.github.io/lean-pool/exposition/}{exposition site} presents
the archive at the level of projects, mathematical results, and supporting
declarations. Each project's card supplies an informal account, attribution, source
references, and links to its headline results. Selecting a result connects this
account to its Lean statement and the surrounding development. This combines the
contributor's explanation of what was formalized with structure extracted from the
checked code. The site builds on the Lean Machine Learning exposition
tools~\citep{degenne2026exposition}.

Table~\ref{tab:exposition} summarizes the coverage of this common interface across
independently developed projects.
\begin{table}[htbp]
\centering\small\setlength{\tabcolsep}{4pt}\renewcommand{\arraystretch}{1.08}
\caption{Coverage of the deployed Exposition export. Nodes are source-visible declarations and edges are intra-project dependencies, with generated auxiliaries followed internally. The scaffold is excluded. The deployment precedes the latest import batch. Its covered projects and declaration and dependency counts refer to the export revision specified in the observation table.}
\label{tab:exposition}
\begin{tabular}{lr}
\toprule
Exposition coverage & Count\\
\midrule
Documented projects & 203\\
Source declarations & 173,362\\
Theorems and lemmas & 124,412\\
Dependency links & 1,389,967\\
Declarations used by multiple others & 87,408\\
\bottomrule
\end{tabular}
\end{table}

The dependency graph exposes the supporting definitions and lemmas behind a
result. Readers can follow its prerequisites or inspect the declarations that use
it, while highlighted headline results provide entry points into a larger
development. The declaration viewer also exposes direct and transitive dependency
counts, allowing readers to locate results with substantial supporting developments
and intermediate declarations shared by later proofs. These are dependencies
within the formalized project; the connections between papers and between imported
projects are a separate level of organization.

Table~\ref{tab:exposition-projects} illustrates the range of project structures
already available through this interface.
\begin{table}[htbp]
\centering\small\setlength{\tabcolsep}{4pt}\renewcommand{\arraystretch}{1.08}
\caption{Examples of the mathematical structures exposed by Exposition. Counts are recomputed from the retained project graphs. Large developments expose reusable intermediate results as well as their headline theorems; graph size measures formal dependency structure, not mathematical importance.}
\label{tab:exposition-projects}
\begin{tabular}{lrr}
\toprule
Project & Declarations & Dependencies\\
\midrule
G{\"o}del's incompleteness theorems & 7,522 & 137,739\\
Classification of compact surfaces & 7,833 & 79,052\\
Navier--Stokes and Euler & 48,626 & 530,547\\
Language generation in the limit & 419 & 1,823\\
\bottomrule
\end{tabular}
\end{table}

Figure~\ref{fig:exposition-overview} shows the Incompleteness project; a selected
theorem's statement panel appears in Appendix~\ref{app:exposition}.
\begin{figure}[t]
\centering\includegraphics[width=\linewidth]{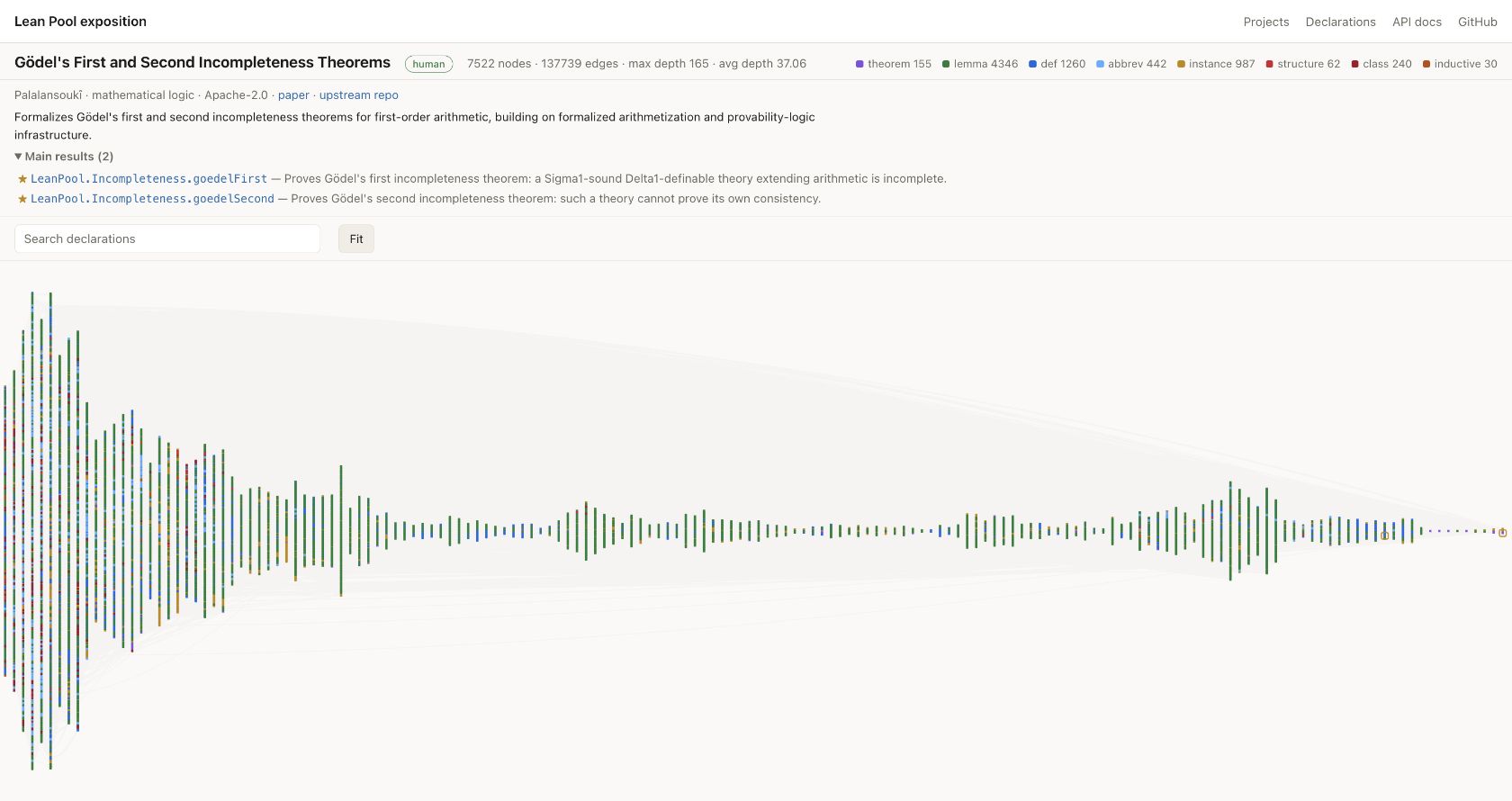}
\caption{The Incompleteness project in Exposition. The project card identifies its
formalization author and upstream source and lists the headline results. The graph
places declarations within the supporting development, making its structure visible
alongside the informal account. The headline results include G\"odel's incompleteness
theorems, while the surrounding graph exposes the formal infrastructure on which
they depend. Selecting a declaration links its position in the graph to its
statement, source, and API documentation. This makes the project accessible both
from its main mathematical claims and from the lemmas used to establish them. The screenshot and coverage tables
use the deployment identified in Appendix~\ref{app:methods}.}
\label{fig:exposition-overview}
\end{figure}

The graph and statement panels complement the source and API documentation:
the graph identifies relevant declarations, and the linked documentation provides
their full formal context. In particular, a reader can move from a project's
informal claim to the assumptions and definitions used in its formal statement
before building on the result. Section~\ref{sec:retrieval} places this inspection
step within the archive's discovery and reuse workflow.

Keeping this view current is part of maintaining the library. The documentation
pipeline checks the correspondence between exported results and project cards.
Recent changes reuse unchanged project graphs and successful CI build outputs when
publishing documentation, avoiding repeated extraction and compilation while keeping
verification of the published data. Appendix~\ref{app:exposition} describes the
conditions under which those outputs can be reused.

\FloatBarrier
\section{Finding and building on archived results}\label{sec:retrieval}
The preferred discovery tools are Octo semantic search, the exposition site, and
project cards, connected by the workflow in Table~\ref{tab:discovery}.
\begin{table}[htbp]
\centering\small\renewcommand{\arraystretch}{1.10}
\caption{Finding a result and following it into a development. A mathematical query,
subject, or known paper provides an entry point. The reader can then inspect the
candidate's assumptions and definitions before importing it alongside other archived
work in the shared environment.}
\label{tab:discovery}
\begin{tabular}{L{.20\linewidth}L{.32\linewidth}L{.39\linewidth}}
\toprule
Starting point & Preferred tool & What the reader obtains\\
\midrule
A mathematical question & \href{https://octo.axiomatic-ai.com/search?scopes=repo\%3AVilin97\%2Flean-pool}{Octo semantic search} & Candidate declarations across the archive\\
A subject or project & \href{https://vilin97.github.io/lean-pool/exposition/}{Exposition} & Project descriptions, statements, and dependency graphs\\
A known paper or result & Project card & Attribution, source references, and formal headline results\\
A candidate to reuse & \href{https://vilin97.github.io/lean-pool/}{Source and API documentation} & Full assumptions, supporting definitions, and import paths\\
\bottomrule
\end{tabular}
\end{table}

\section{Related work}\label{sec:discussion}
\paragraph{Archives and shared libraries.}
AFP combines contributed formalizations with review and continuing
maintenance~\citep{afp2026,afpsubmission2026,mackenzie2021afpevaluation}.
Mathlib develops an integrated mathematical library~\citep{mathlib2020,baanen2026growing}.
Reservoir indexes Lean packages, the Rocq Platform distributes compatible packages,
and Software Heritage preserves source histories~\citep{reservoir2026,rocqplatform2026,abramatic2018heritage}.
Table~\ref{tab:library-comparison} compares Lean Pool with Tau Ceti, Palomar Registry,
and Mathlib along their contribution, maintenance, acceptance, and documentation policies.
\begin{table}[!t]
\centering\footnotesize\setlength{\tabcolsep}{3pt}\renewcommand{\arraystretch}{1.12}
\caption{Contribution and maintenance models in the Lean ecosystem, as documented
by the projects~\citep{poolci2026,tauceti2026,palomarpolicy2026,mathlibcontribute2026,mathlib2020}.
Lean Pool maintains independent developments together. Tau Ceti and Mathlib integrate
contributions into a shared mathematical API, with different roles for human and AI
contributors. Palomar registers checked versions of separate repositories; updated
versions require resubmission, verification, and review.}
\label{tab:library-comparison}
\begin{tabular}{L{.12\linewidth}L{.19\linewidth}L{.18\linewidth}L{.23\linewidth}L{.18\linewidth}}
\toprule
Project & Contributions & Maintenance & Acceptance criteria & Documentation\\
\midrule
Lean Pool & Human, AI, and mixed projects; direct submissions and attributed imports
& Agents propose upgrades and optimizations; maintainers oversee merges
& Completed proofs; permitted axioms; attributed cards; permissive license; CI and mathematical review
& Project cards, informal results, dependency graphs, source-linked API\\\addlinespace
Tau Ceti & AI-authored mathematics following human roadmaps
& AI workers prioritize maintenance and review, update Mathlib pins, and repair the library
& CI and adversarial AI review; roadmap fit, reuse, API coherence, and compatibility with Mathlib
& Human roadmaps, module documentation, declaration docstrings\\\addlinespace
Palomar Registry & Human, AI, or mixed formalizations submitted by an author or authorized maintainer
& Verified versions persist; authors resubmit updates
& Lean and NanoDa proof checks; automated review of statement faithfulness, disclosure, and research interest
& Recorded statements, provenance, dependencies, and public verification and review reports\\\addlinespace
Mathlib & Human contributors; disclosed AI use with contributor understanding and justification
& Community maintenance across Lean releases
& Human review, generality, integration, maintainability, style, and CI
& Module and declaration documentation, API reference, library overviews\\
\bottomrule
\end{tabular}
\end{table}

\paragraph{Discovery, formalization, and reuse.}
TheoremSearch retrieves statements from mathematical literature, while TheoremGraph
connects statements and their dependencies across informal and formal
sources~\citep{alexander2026semantic,kurgan2026theoremgraph}.
LeanDojo and LeanAgent study retrieval and learning for proof
construction~\citep{yang2023leandojo,kumarappan2025leanagent}.
Project-level benchmarks evaluate reasoning in existing libraries and software
contexts~\citep{hu2025minictx,xin2026verisoftbench,ye2026vero,klingner2026evaluation}.
Semi-autonomous formalization connects informal arguments to checked
developments~\citep{ilin2026vml}. The LeanEval structural audit studies the resulting
code and documents cross-project reuse~\citep{leanevalanalysis2026}.

\paragraph{Repair and mathematical review.}
Compatibility studies, proof transport, and APRIL's compiler-feedback repair address
the maintenance of formal proofs~\citep{luan2025compatibility,ringer2021repair,wang2026april}.
Statement-evaluation methods examine correspondence between informal claims and formal
expressions~\citep{poiroux2025reliable,liu2025beq,lu2025formalign,zhang2026beyondcompilation,ilin2026equivalence}.
Expert assessments of generated mathematics identify obligations beyond filling
proof gaps, including appropriate definitions, faithful statements, and usable library
design~\citep{ilin2026sorries,meek2026numerical}.

\section{Conclusion}
\pool{} brings completed formalizations into a common environment maintained through
agent-assisted upgrades, optimization, review, and community contribution. Its
operational history documents repeated maintenance, while the LeanEval audit shows
reuse in subsequent research. As more mathematical papers acquire formal proofs,
the archive provides a place to preserve their attribution and keep their dependencies
usable for later work.

\clearpage
\label{firststatementpage}
\subsection*{AI use statement}
Generative AI assisted the collection and organization of public-source evidence,
analysis design and implementation, interpretation of operational records,
literature discovery, figure generation, and drafting and revision of this paper.
The archive's agents and review models are themselves objects of the study and are
described separately in the paper. Reported numerical results are computed from
retained source and execution records; they are not synthetic measurements.
Validation checks source hashes, recomputes reported aggregates, and reconstructs
the tables and figures. The author takes responsibility for the final manuscript.

\subsection*{Reproducibility statement}
Appendix~\ref{app:methods} defines the observation periods and measurement
populations. The tables identify source revisions and link the public PR reports
underlying the historical analyses. Measurements executed for this paper are
distinguished from those reports. Raw records and analysis programs are retained
separately from the manuscript submission.

\section*{Acknowledgments}
We thank Justin Asher for co-creating Lean Pool and contributing its initial
discovery tooling and continuous integration, and Austin Letson for contributing
proof golfing. We thank the archive's contributors and the authors of
the imported formalizations for making their work available to the community.

\clearpage
\ifpaperlocal
  \bibliographystyle{style/iclr2027_conference}
  \bibliography{../references}
\else
  \bibliographystyle{paper/style/iclr2027_conference}
  \bibliography{references}
\fi
\clearpage
\appendix
\section{Observation scope and measurement methods}\label{app:methods}
\label{app:presentation-methods}
The archive and its automation repository continue to change. Table~\ref{tab:observations} identifies the observations analyzed in this paper. Tables identify source revisions and link the original PR reports. Raw
measurements and analysis inputs are retained separately.
The full project/publication concordance follows in Appendix~\ref{app:project-papers}.

\begin{table}[h]
\centering\small\renewcommand{\arraystretch}{1.10}
\caption{Observation periods and populations. The archive, its automation repository, and the exposition site continue to change; these dates specify the source observations and measurements used here.}
\label{tab:observations}
\begin{tabular}{L{.29\linewidth}L{.60\linewidth}}
\toprule
Evidence & Observation\\
\midrule
Archive source census & September 21, 2026, 06:07 UTC\\
Public PR records & Retrieved September 21, 2026\\
Structured review-service reports & Last retained report: September 13, 2026\\
Exposition export & September 20, 2026, 14:24 UTC\\
LeanEval reuse analysis & Analysis retrieved September 12, 2026\\
Automation job definitions & Unchanged revision rechecked September 21, 2026\\
Build resources & Clean builds measured September 21, 2026 (UTC)\\
Optimization comparison & Historical revisions rebuilt September 13--14, 2026 (UTC)\\
Exposition screenshots & Live interface captured September 21, 2026 (UTC)\\
\bottomrule
\end{tabular}
\end{table}

The source census counts physical lines in the completed library, including comments,
blank lines, and its internal scaffold, and excludes dependency source and the
generated top-level import index. Commit contributors and merged-PR authors are
counted separately using GitHub accounts classified as users, excluding bots.
First imports are identified from
merge commits against their first parents, avoiding duplicate counts from stacked PRs.

The declaration-command census counts source syntax, including explicitly public
declarations, and excludes examples and compiler-generated auxiliaries. The exposition census counts declarations with
source locations and follows auxiliaries when resolving dependencies; it uses a
coherent deployed export that precedes the latest import batch. These populations
are reported separately. The stable-version migration passed the native
combined-library CI build and repository checks.

\paragraph{Research records.}\label{app:retained-evaluations}

The analysis uses retained source records, build logs and memory samples, review
reports, and project and publication records. Validation checks file hashes and
regenerates the reported displays without compiler or model calls.

The research uses model assistance for collection, analysis code, mathematical
inspection, and writing. Review-service outputs are identified as model-generated
reports. Public-source attribution and the full project bibliography are retained
independently of those assessments.

\subsection{Composition of the collection}
Table~\ref{tab:project-scale} characterizes the largest developments in the archive.
\begin{table}[htbp]
\centering\small\setlength{\tabcolsep}{4pt}\renewcommand{\arraystretch}{1.08}
\caption{Largest archived projects by physical source size. Counts include comments, blank lines, and integration changes within each project. Provenance comes from the attributed project card. The examples span analysis, geometry, logic, number theory, and computer science; source size describes the formal development rather than the mathematical significance of its headline result. The full catalogue links every project to its upstream source and publications.}
\label{tab:project-scale}
\begin{tabular}{L{.66\linewidth}lr}
\toprule
Project & Provenance & Lean lines\\
\midrule
Finite-time breakdown for Navier--Stokes and Euler & AI & 641,073\\
A complex structure on the six-sphere & AI & 260,646\\
Classification of Compact Surfaces & AI & 143,423\\
Polynomial-factor hardness of the closest vector problem & mix & 135,818\\
Improved asymptotic bounds for binary and spherical codes & mix & 112,078\\
The Convex-Octagon Case of Erd{\H{o}}s Problem 97 & AI & 101,724\\
Jordan--Sch{\"o}nflies theorem & AI & 77,482\\
Quantum parallel repetition & mix & 71,975\\
Modular forms and the generalized residue theorem & mix & 67,709\\
Ehrhart's sharp volume inequality & mix & 56,357\\
Sharp asymptotic upper bounds for sphere packing & mix & 55,713\\
G{\"o}del's First and Second Incompleteness Theorems & human & 55,430\\
\bottomrule
\end{tabular}
\end{table}

The archive retired the forward-Euler and special-numbers projects through a
separate curation PR. Its final record cites the archive's project-size threshold
and the textbook scope of the material, respectively. This change is separate from
the compiler repairs and is labeled as project retirement in the growth figure.

\subsection{Build-resource comparison}\label{app:build-methods}
Tables~\ref{tab:build-resources}, \ref{tab:build-configuration}, and~\ref{tab:build-details}
report clean builds of the current archive and the comparison libraries, together
with their configuration and exact source revisions.
\begin{table}[htbp]
\centering\small\setlength{\tabcolsep}{4pt}\renewcommand{\arraystretch}{1.10}
\caption{Clean builds of the named libraries on the same Azure VM. Lean Pool has more source lines and a longer elapsed build than the matching Mathlib release in these runs. Lean Pool retains its fetched Mathlib cache; Mathlib rebuilds its own sources with its external dependencies prepared. The matching-release comparison uses the same compiler and Mathlib revision. Current Mathlib and Tau Ceti use their supported compiler versions, listed in Appendix~\ref{app:build-methods}. Source size excludes dependency code. CPU time sums user and system time; RAM is sampled peak combined proportional resident memory, accounting for shared pages. Each row is one complete build on a shared VM.}
\label{tab:build-resources}
\begin{tabular}{lrrrr}
\toprule
Library & Source MLOC & Build min $\downarrow$ & CPU h $\downarrow$ & RAM GiB $\downarrow$\\
\midrule
Lean Pool & 3.23 & 60.28 & 14.29 & 20.03\\
Mathlib (matching release) & 2.33 & 37.44 & 9.38 & 7.35\\
Mathlib (current) & 2.33 & 35.83 & 8.99 & 7.00\\
Tau Ceti & 1.46 & 30.89 & 7.13 & 10.67\\
\bottomrule
\end{tabular}
\end{table}

\begin{table}[htbp]
\centering\small\setlength{\tabcolsep}{4pt}\renewcommand{\arraystretch}{1.10}
\caption{Shared measurement configuration. Builds run sequentially with the same CPU affinity and thread settings. The observer verifies that every source module compiles, its output exists, and the source checkout remains unchanged. Memory peaks are sampled between process scans. Cache downloads, documentation, and separate CI audits are outside the timed command. Normal in-build lint options remain enabled. The VM ran background tasks during these measurements; their activity is recorded separately and is outside the reported CPU and process-memory totals.}
\label{tab:build-configuration}
\begin{tabular}{lL{.64\linewidth}}
\toprule
Setting & Value\\
\midrule
Host & Azure Linux VM; AMD EPYC 9V45 processor\\
Available CPUs / Lean threads & 16 / 16\\
Physical RAM & 125.8 GiB\\
Dependency preparation & Prebuilt; outside timing\\
Root-library outputs & Absent before each build\\
Filesystem cache & Compilation inputs preloaded before timing\\
Artifact restoration & Disabled for the measured root library\\
Run order & Lean Pool, matching Mathlib, current Mathlib, Tau Ceti\\
Time instrumentation & GNU time; elapsed and aggregate user + system CPU\\
Memory instrumentation & Summed proportional resident memory of the build process group\\
\bottomrule
\end{tabular}
\end{table}

\begin{table}[htbp]
\centering\small\setlength{\tabcolsep}{4pt}\renewcommand{\arraystretch}{1.10}
\caption{Measured source revisions and compiler versions. Links identify the exact source for each completed build. The matching Mathlib release is the dependency pinned by Lean Pool. The current Mathlib observation uses a newer compiler; Tau Ceti uses its own supported release candidate.}
\label{tab:build-details}
\begin{tabular}{lll}
\toprule
Library & Lean & Source revision\\
\midrule
Lean Pool & 4.34.0 & \href{https://github.com/Vilin97/lean-pool/commit/bc6f18b24cf19f89bb117189e75cc11efea1edf6}{bc6f18b2}\\
Mathlib (matching release) & 4.34.0 & \href{https://github.com/leanprover-community/mathlib4/commit/5ed2965256430c3649e86755f9576b54eca72435}{5ed29652}\\
Mathlib (current) & 4.35.0-rc2 & \href{https://github.com/leanprover-community/mathlib4/commit/0a6c8e0355da0405d616f80b9f8232c4fab2cc5b}{0a6c8e03}\\
Tau Ceti & 4.34.0-rc2 & \href{https://github.com/TauCetiProject/TauCeti/commit/8aa2ad025fa7c4c6159b8da515b75cf1bebcbcd6}{8aa2ad02}\\
\bottomrule
\end{tabular}
\end{table}

\section{CI and profiling}\label{app:ci}
\pool{} builds the combined library, rejects unexpected warnings, checks generated
project indexes, runs Mathlib's environment and text-style linters, and applies its
repository-specific quality checker. Cold CI builds can be split across project
shards; the assembled outputs are followed by a combined-library build and common
checks. Python tooling has formatting, linting, and unit-test checks, while workflow
checks validate the CI definitions.

The environment linters check simplifier normal forms, unused arguments, declaration
types, and structures whose fields are all propositions. Source linters check naming,
headers, whitespace, and layout conventions. Tau Ceti records existing exceptions in
a baseline and rejects new violations; Pool rejects linter waivers in project content.
Tau Ceti also checks duplicate declaration ownership, including orphan modules,
before imported environments can hide collisions.

\begin{table}[t]
\centering\small\renewcommand{\arraystretch}{1.10}
\setlength{\tabcolsep}{4pt}
\caption{Mechanical checks and performance tooling in the observed repositories,
supported by the retained workflow definitions and benchmark
documentation~\citep{poolci2026,mathlibci2026,taucetici2026}. All use library compilation
and Mathlib linters. Lean Pool adds project-level admission checks, while Tau Ceti
requires a performance comparison for merging; Pool profiling is advisory. Mathlib
maintains whole-build benchmark and telemetry infrastructure.}
\label{tab:ci-comparison}
\begin{tabular}{L{.19\linewidth}L{.24\linewidth}L{.24\linewidth}L{.23\linewidth}}
\toprule
Responsibility & Lean Pool & Mathlib & Tau Ceti\\
\midrule
Compilation & Combined archive build; warning rejection & Library build, tests, warning checks & Complete source-tree build; warning and output checks\\
Declaration lint & Mathlib environment linters & Mathlib lint driver & Mathlib environment linters with a recorded baseline\\
Source style & Mathlib style checks; no waivers & Source and bibliography style checks & Style and header checks; bounded files\\
Logical assumptions & Allowed-axiom check and compiled-environment audit & Standard trusted Lean foundations; library review and linting & Allowed-axiom audit of compiled sources\\
Project governance & Attributed cards, completed results, permissive license, separated content PRs & Review of shared mathematical API & AI-contributed mathematics under trusted governance checks\\
Resource policy & No content option overrides; source and proof size limits & Library-configured options and reviewed exceptions & No content option overrides; per-process watchdog\\
Performance & Advisory per-file profiling & Build benchmark and telemetry tooling & Required comparison using hardware instructions or CPU time; advisory heartbeats\\
\bottomrule
\end{tabular}
\end{table}

\paragraph{Archive-specific rules.}
Cards must identify authors, a primary source, proof provenance, subject tags, and
formal main results paired with informal statements. The source license must be
\code{Apache-2.0} or \code{MIT}. Registered results must resolve in Lean, and generated
cards must agree with the registry. Completed projects cannot contain unfinished
proofs, new axioms, unchecked declarations, linter waivers, diagnostic commands, or
broad imports of all Mathlib. The environment audit also checks generated declarations
for forbidden assumptions and programmatic attempts to change restricted options.

The separate challenge board permits explicit open statements. Their eventual
solutions are checked against the registered statements with an independent kernel
comparison. This exception does not apply to completed projects.

\paragraph{Profiling.}
For new files, the Pool profiler reports absolute source size, declaration counts,
heartbeats, and elapsed elaboration measurements. For modified files it reports a
before/after comparison and a statement-change summary. Heartbeats measure Lean's
internal allocation-based computation counter. They complement elapsed time but do
not replace a whole-build measurement. The displayed comment can summarize the
largest file changes; raw measurements are retained as workflow artifacts.

Mathlib's build benchmark records whole-build instructions, CPU and elapsed time,
peak memory, per-module measurements, and critical build paths. Tau Ceti's performance
workflow compares immutable base and head revisions using retired instructions when
available and CPU time otherwise. Its trusted measurement harness keeps candidate
code separate from the recorded counters. Pool profiling is advisory in the observed
configuration, whereas Tau Ceti publishes a required performance status.
Table~\ref{tab:ci-comparison} compares the responsibilities of these workflows.

\section{Daily jobs and contributor participation}\label{app:daily-jobs}
The separate Lean Pool automation repository contains maintained prompts and a host
runner. The runner refreshes its checkout before execution and prevents duplicate
instances of the same job. A configured fallback backend can continue a job after an
execution failure. Job definitions specify responsibilities; they do not establish
which model performed each historical contribution.

\begin{table}[h]
\centering\small\renewcommand{\arraystretch}{1.10}
\caption{Daily responsibilities in the maintained automation repository. Discovery
jobs prepare attributed imports; review and issue jobs act on existing submissions;
optimization jobs improve archived code; announcements connect accepted results to
the community. These are job responsibilities, not a census of completed executions.
Dependency bumping runs in a separate scheduled repository workflow.}
\label{tab:daily-jobs}
\begin{tabular}{L{.25\linewidth}L{.64\linewidth}}
\toprule
Job & Work performed\\
\midrule
Recent discovery & Find recently completed formalizations; check attribution, license, fit, dependencies, and cost; prepare an import PR.\\
Older discovery & Search the preceding literature and repositories for completed projects missed by recent discovery; apply the same admission checks.\\
Open-PR review & Inspect current submissions, mathematical references, CI, reviews, and profiling; repair eligible branches or record unresolved issues.\\
Project optimization & Profile an existing project, shorten proofs or reduce compilation cost, validate the change, and open a PR.\\
Maintainer issues & Inspect submissions awaiting a maintainer decision; resolve eligible source or metadata problems and retain unresolved decisions.\\
Announcements & Identify newly accepted projects, check their advertised results, and announce them with attribution while suppressing duplicate posts.\\
\bottomrule
\end{tabular}
\end{table}

Dependency bumping is a separate scheduled repository workflow. It detects a release,
prepares the updated environment, probes projects, assigns failures to repair agents,
and assembles their patches. It then builds the combined archive and creates a draft
PR reporting the outcome. Optimization and repair proposals remain subject to the
repository checks and maintainer decisions.

\subsection{What other contributors have added}\label{app:community}
Table~\ref{tab:community} attributes further contributions to the archive.
\begin{table}[htbp]
\centering\small\setlength{\tabcolsep}{4pt}\renewcommand{\arraystretch}{1.08}
\caption{Contributions from community submitters through the current observation. Counts cover merged PRs; descriptions summarize the mathematics and tooling in their accepted changes. Project cards separately preserve upstream formalization authorship.}
\label{tab:community}
\begin{tabular}{l r L{.58\linewidth}}
\toprule
Contributor & Merged PRs & Contributions\\
\midrule
\href{https://github.com/lyfar}{lyfar} & 13 & Combinatorics, geometry, and project corrections\\
\href{https://github.com/jtraverso}{jtraverso} & 7 & Convex duality, graph decompositions, matching, flows, and concentration\\
\href{https://github.com/Lemmy00}{Lemmy00} & 2 & Pythagorean parametrization; Cramer--Wold\\
\href{https://github.com/ldct}{ldct} & 2 & Commuting probability; Chudnovsky formula\\
\href{https://github.com/AlexeyMilovanov}{AlexeyMilovanov} & 1 & Harper's isoperimetric theorem\\
\href{https://github.com/ElNando888}{ElNando888} & 1 & Krafft sieve\\
\href{https://github.com/Hydrodynamical}{Hydrodynamical} & 1 & Vlasov mean-field limit\\
\href{https://github.com/JonCYeh}{JonCYeh} & 1 & Pentagonal number theorem and analytic Jacobi triple product\\
\href{https://github.com/dillon-11}{dillon-11} & 1 & Lehmer polynomial and Coxeter element\\
\href{https://github.com/gersh}{gersh} & 1 & Three-body nonintegrability\\
\href{https://github.com/jarfo}{jarfo} & 1 & Unique multiset sums\\
\href{https://github.com/mccorvie}{mccorvie} & 1 & Classification of surfaces\\
\href{https://github.com/rkirov}{rkirov} & 1 & Jacobian of a Riemann surface\\
\href{https://github.com/vaguiarl}{vaguiarl} & 1 & Odlyzko bound solution\\
\bottomrule
\end{tabular}
\end{table}

\section{Stable-version migration}\label{app:migration}
Table~\ref{tab:stable-migration} follows the stable-version migration from the original
production probe to the accepted integration.
\begin{table}[htbp]
\centering\small\setlength{\tabcolsep}{4pt}\renewcommand{\arraystretch}{1.08}
\caption{Production record of the stable-version migration. The initial agent jobs produced repair patches; additional agent-assisted integration addressed remaining build errors, warning changes, and API compatibility. Accepted membership reflects intervening imports and the separately approved project retirements. Source churn compares the merge with its first parent and therefore excludes those unrelated changes. Job success is the workflow status, not a guarantee that its patch completes the combined migration.}
\label{tab:stable-migration}
\begin{tabular}{lr}
\toprule
Recorded stage & Count\\
\midrule
Probe projects & 191\\
Compiler failures & 97\\
Warning only projects & 32\\
Repair jobs & 97\\
Repair jobs succeeded & 95\\
Repair jobs failed & 2\\
Accepted projects & 198\\
Changed files & 639\\
Line churn & 5,951\\
\bottomrule
\end{tabular}
\end{table}

The failed initial repair jobs concerned the graph fundamental-group and
incompleteness developments. Subsequent integration repaired the combined source,
resolved warnings, and adapted supporting APIs to changed Mathlib interfaces.
The PR records strengthened auxiliary hypotheses concerning measurability,
sigma-finiteness, and maximality, alongside import and namespace repairs. Its
successful build establishes compatibility of the accepted statements; it does not
establish equivalence of every declaration type before and after the upgrade.
The accepted PR links the migration record and successful CI run.

Table~\ref{tab:migration-effort} reports the changes that landed in each upgrade.
\begin{table}[htbp]
\centering\small\setlength{\tabcolsep}{4pt}\renewcommand{\arraystretch}{1.08}
\caption{Source changes in the accepted dependency upgrades. Churn is added plus removed physical Lean lines. It includes cleanup in projects that already compiled and therefore describes the work that landed rather than the minimum repair. These records provide an observable proxy for maintenance effort; historical monetary charges and human labor were not logged.}
\label{tab:migration-effort}
\begin{tabular}{lrr}
\toprule
Target Lean & Files edited & Lines changed \\
\midrule
4.31.0-rc1 & 825 & 25,583 \\
4.32.0-rc1 & 402 & 5,653 \\
4.33.0-rc1 & 576 & 11,690 \\
4.33.0-rc2 & 585 & 5,221 \\
4.34.0-rc1 & 448 & 4,676 \\
4.34.0 & 639 & 5,951 \\
\bottomrule
\end{tabular}
\end{table}

\section{Accepted optimization changes}\label{app:optimization}
The historical optimizations in Table~\ref{tab:optimization-main} were evaluated against their parent revisions by rebuilding the whole
Lean Pool library on the same Azure VM. External dependencies are prepared before
timing. The library's own build outputs are removed before each run, and dependency
and toolchain inputs are read into memory outside the timed command. The dependency
manifest and build configuration are identical within each comparison. Runs are
sequential with the same CPU affinity and Lean thread settings.

Table~\ref{tab:optimization-details} gives the complete build measurements.
\begin{table}[htbp]
\centering\small\setlength{\tabcolsep}{4pt}\renewcommand{\arraystretch}{1.10}
\caption{Complete-library builds underlying the optimization comparison. Each row reports one clean run, with the parent before the accepted change. The VM uses the same CPU affinity and Lean thread setting as the library comparison. Dependencies are prebuilt and their compilation inputs preloaded outside timing. CPU hours sum user and system time across the build; RAM is combined proportional resident memory sampled throughout the run. Paired measurements use each PR's own source and toolchain, so comparisons are within PRs rather than between different archive sizes.}
\label{tab:optimization-details}
\begin{tabular}{lllrrr}
\toprule
PR & Revision & Lean & Wall min $\downarrow$ & CPU hours $\downarrow$ & RAM GiB $\downarrow$\\
\midrule
\href{https://github.com/Vilin97/lean-pool/pull/117}{117} & Before & 4.30.0-rc2 & 5.30 & 0.96 & 15.60\\
 & After & 4.30.0-rc2 & 5.25 & 0.94 & 15.57\\
\href{https://github.com/Vilin97/lean-pool/pull/187}{187} & Before & 4.32.0-rc1 & 16.11 & 4.04 & 26.12\\
 & After & 4.32.0-rc1 & 15.59 & 3.90 & 26.85\\
\href{https://github.com/Vilin97/lean-pool/pull/246}{246} & Before & 4.32.0-rc1 & 21.83 & 5.51 & 25.95\\
 & After & 4.32.0-rc1 & 23.01 & 5.81 & 24.58\\
\href{https://github.com/Vilin97/lean-pool/pull/338}{338} & Before & 4.33.0-rc2 & 30.22 & 7.66 & 21.02\\
 & After & 4.33.0-rc2 & 29.13 & 7.38 & 20.89\\
\href{https://github.com/Vilin97/lean-pool/pull/339}{339} & Before & 4.34.0-rc1 & 28.46 & 7.20 & 19.81\\
 & After & 4.34.0-rc1 & 26.82 & 6.78 & 19.55\\
\bottomrule
\end{tabular}
\end{table}

The measurement script checks that every source module was compiled and that its
compiled output exists. It samples the proportional resident memory of the build's
process group, counting shared pages proportionally across workers. The same concurrent-memory measure is used
for the comparison with Mathlib and Tau Ceti. Source changes include refactoring and deletion of
existing declarations as well as shortening proof bodies.

The project-level comparisons in Table~\ref{tab:recent-optimizations} retain their
original measurement scopes. Restricted-sum and interior-point results use medians of repeated clean project
builds with warm dependencies. Infinite Connes rigidity compares a single baseline
with the median of repeated optimized builds. The Burkholder and quantum parallel
repetition results each use a single clean build per revision on the same host. The fluid-equation refactor uses an interleaved
before/after ordering on Azure and adds APIs for downstream reuse. Its measured
whole-project compilation time does not decrease, although the retained downstream
client comparison uses fewer instructions and less memory. Tactic-import cleanup
has differing cache preparation between runs; its source reduction is reported
without a timing comparison. These observations are not pooled with the controlled
whole-library measurements.

\paragraph{Library-wide module and proof optimization.}
Table~\ref{tab:module-optimization} records the historical benchmark supporting
the merged library-wide change; its integrated version is included in the current
source census and build-resource comparison.
\begin{table}[htbp]
\centering\small\setlength{\tabcolsep}{4pt}\renewcommand{\arraystretch}{1.08}
\caption{Fixed-workload benchmark accompanying the accepted module and proof optimization. The \href{https://github.com/Vilin97/lean-pool/pull/415}{merged PR} compares the same original archive workload with pinned dependencies and cold project caches on a shared WSL host. Its measured candidate precedes later imports and final integration. Retired instructions decrease alongside memory use in this historical comparison. Process RSS is the largest individual process; anonymous memory sums sampled concurrent private allocations and excludes file-backed mappings. These differ from the proportional-memory metric in the fresh Azure comparison. The current source census and Azure build include the merged optimization.}
\label{tab:module-optimization}
\begin{tabular}{lrrr}
\toprule
Workload & Instructions T $\downarrow$ & Process GiB $\downarrow$ & Anonymous GiB $\downarrow$\\
\midrule
Original source & 124.499 & 12.605 & 13.222\\
Optimized source & 103.172 & 2.891 & 5.594\\
\bottomrule
\end{tabular}
\end{table}

\section{Production reviews and agreement}\label{app:reviews}
Table~\ref{tab:review-agreement} describes agreement within the retained histories
of repeatedly reviewed PRs.
\begin{table}[htbp]
\centering\small\setlength{\tabcolsep}{4pt}\renewcommand{\arraystretch}{1.08}
\caption{Agreement between retained reports of the same review type. Same-PR comparisons can span changed code and review models. Same-revision comparisons require an explicit matching reviewed commit. Report pairs are dependent when a PR has repeated reviews; agreement describes verdict consistency rather than mathematical correctness.}
\label{tab:review-agreement}
\begin{tabular}{lrr}
\toprule
Comparison & Pairs & Same verdict\\
\midrule
Same PR & 69 & 37\\
Same reviewed revision & 0 & 0\\
\bottomrule
\end{tabular}
\end{table}

Table~\ref{tab:review-coverage} separates the report types and explicit coverage limitations.
\begin{table}[htbp]
\centering\small\setlength{\tabcolsep}{4pt}\renewcommand{\arraystretch}{1.08}
\caption{Composition of the retained review-service reports. Project, challenge, solution, and refactor reviews follow distinct rubrics. Explicit partial-coverage notices are counted separately and may occur in any category.}
\label{tab:review-coverage}
\begin{tabular}{lr}
\toprule
Report category & Reports\\
\midrule
Project & 278\\
Challenge & 7\\
Solution & 2\\
Refactor & 9\\
Explicitly partial reports & 14\\
\bottomrule
\end{tabular}
\end{table}

\paragraph{Service evolution and state.}
The original service used API calls with bounded review input. Its Azure
configuration uses Codex account quota and partitions large changes into source
portions before integrating their findings. The integration stage tracks open
questions and requests original source excerpts when needed. A coverage manifest
records which source ranges were presented. This describes coverage of the input,
not an independent assessment of the model's mathematical accuracy.
The workflow was manually disabled before the current observation. Its retained
reports therefore describe past executions; they do not establish ongoing coverage
of the most recent imports. Free-form daily-agent review comments remain outside
the structured report census.

\paragraph{Price accounting.}
Historical comments expose API-based price estimates. The Azure comments expose
nominal Standard-API-equivalent estimates while execution consumes account quota.
The latter value input at uncached rates and omit VM costs. They are not invoices.
The census retains changed versions of a review when available and deduplicates
identical report revisions. Overwritten or unrecorded executions are not recoverable
from the retained comments, so totals are prices attached to retained reports rather
than the complete expenditure of operating the archive.

\section{Exposition, discovery, and reuse}\label{app:exposition}
\begin{figure}[htbp]
\centering\includegraphics[width=.62\linewidth,trim=1280bp 0bp 0bp 290bp,clip]{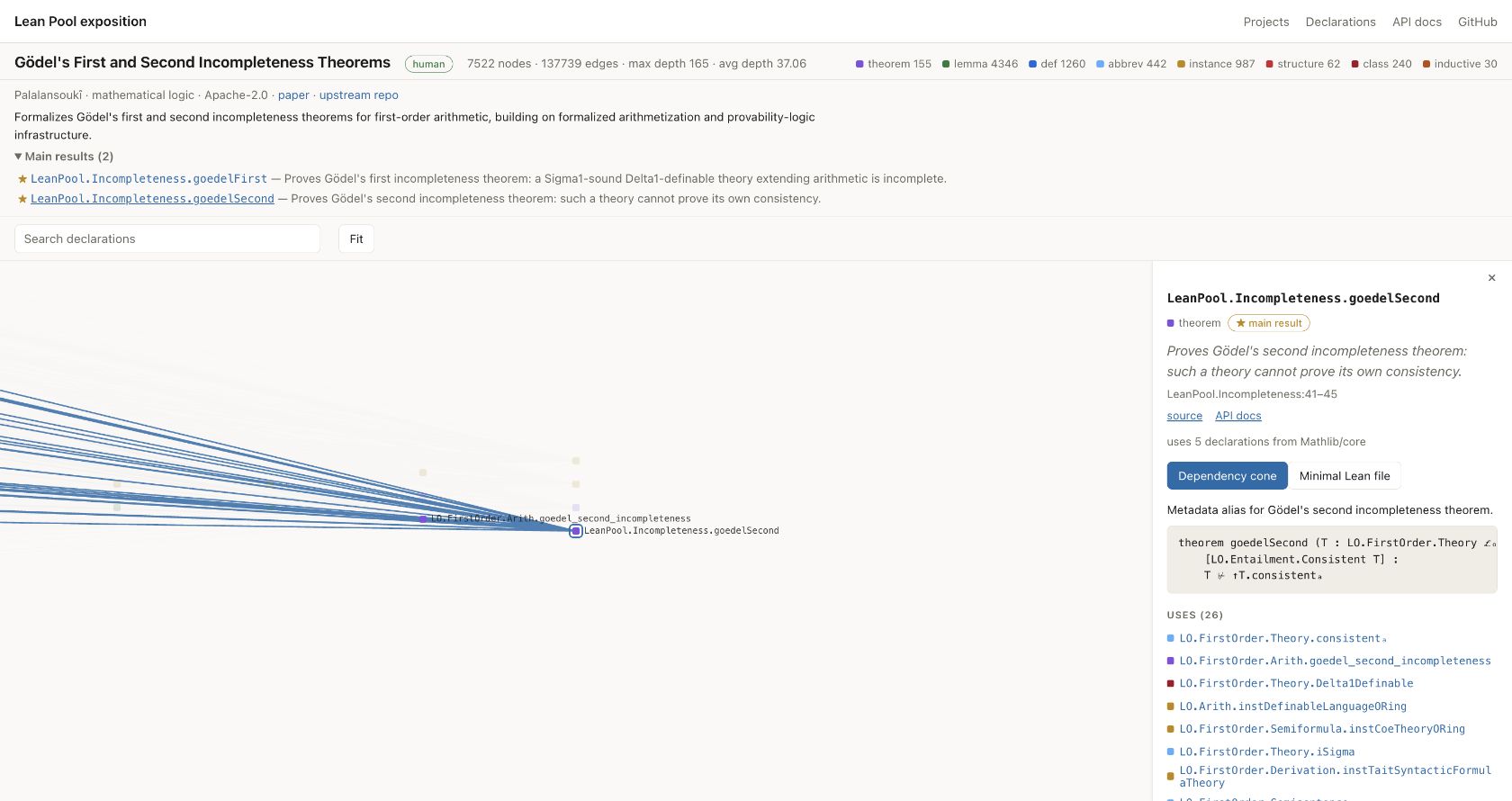}
\caption{The theorem panel for G\"odel's second incompleteness theorem in Exposition.
Selecting a headline result opens its informal description, Lean statement preview, and links
to source and API documentation. This close-up shows how the site connects a result
in the overview graph to its formal account. Capture dates appear in
Table~\ref{tab:observations}.}
\label{fig:exposition-theorem}
\end{figure}

Figure~\ref{fig:exposition-theorem} follows a main result from the project overview
into its formal statement and supporting declarations.

\paragraph{Maintaining the publication pipeline.}
The documentation pipeline caches project graph exports by their source,
transitive import dependencies, toolchain configuration, and extractor version.
A changed dependency invalidates the affected export. Documentation can reuse
successful native-CI Lean artifacts only after matching the source revision and
verified source tree; otherwise it takes the ordinary build path. Data validation
still runs when a cached graph is reused. 

\subsection{Recorded reuse}\label{app:reuse}
Table~\ref{tab:reuse-cases} lists source-level reuse documented in the retained histories.
\begin{table}[htbp]
\centering\small\setlength{\tabcolsep}{4pt}\renewcommand{\arraystretch}{1.08}
\caption{Recorded reuse within and beyond the archive. The Kurosh development applies the existing finite-graph formalization's spanning-tree basis in its Schreier index formula. Other entries distinguish attributed source copying, an explicit package dependency, and use within later projects controlled by the maintainer. These source histories complement the LeanEval declaration-matching analysis.}
\label{tab:reuse-cases}
\begin{tabular}{L{.35\linewidth}L{.53\linewidth}}
\toprule
Consumer & Recorded reuse \\
\midrule
\href{https://github.com/Vilin97/lean-pool/blob/bc6f18b24cf19f89bb117189e75cc11efea1edf6/LeanPool/Kurosh/IndexFormula.lean\#L179}{Kurosh subgroup development} & Applies the archived finite-graph spanning-tree basis \\
\href{https://github.com/pengzhang91/Feige/blob/98ab466e74280ae9d40622c19dc7f24f01b60864/NOTICE.md#L25}{Feige development} & Copies attributed Pool isoperimetry; applies its theorem \\
\href{https://github.com/gersh/poincare-chapter-vi/blob/0cb686c27f141ab52b521715f64cd57da1a62fc9/PoincareChapterVI/ClassicalLeanPool.lean#L25}{Poincar\'e development} & Pins Pool and reexports its nonintegrability theorem \\
\href{https://github.com/Vilin97/homotopy-groups-lean/blob/c66523531ff172d7f41913d94e56921e790a1b47/examples/submissions/sphere_lower_homotopy_subsingleton/Submission/WhiteheadTheorem/Defs.lean#L6}{Maintainer homotopy development} & Contains attributed homotopy modules imported through Pool \\
\href{https://github.com/Vilin97/MazurTheorem/blob/9327963d4ec14fba49c7b14b004fd00707ffc2e9/MazurTorsion/Upstream/CurveCohomologyGrothendieckVanishing.lean#L15}{Maintainer Mazur development} & Uses the archived Grothendieck-vanishing result \\
\bottomrule
\end{tabular}
\end{table}

\begin{figure}[htbp]
\centering\includegraphics[width=\linewidth]{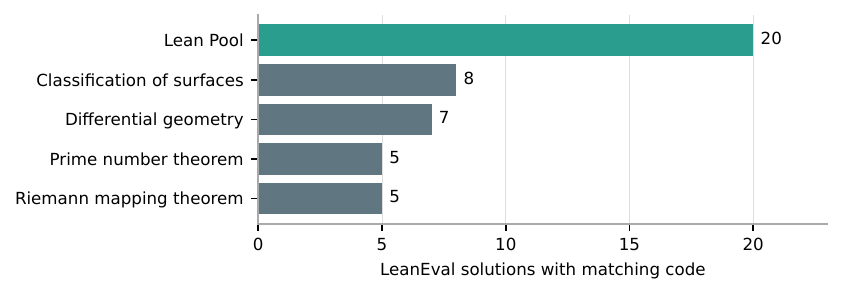}
\caption{Reuse of public repositories in the LeanEval audit~\citep{leanevalanalysis2026}.
Lean Pool has matching declarations in more solutions than any other repository in
the comparison. Counts are solutions with matching normalized source declarations,
including copied code; ordinary package imports such as Mathlib dependencies are
excluded. A solution can match both an archived project and its upstream repository.
The audit's agent traces document how archived mathematics entered subsequent
research-level formalizations. Further source-level reuse appears in
Appendix~\ref{app:reuse}.}
\label{fig:leaneval-reuse}
\end{figure}

\section{Original upstream Lean versions}\label{app:source-versions}
Table~\ref{tab:source-versions} records the upstream environment for each imported project.
\begingroup\small\setlength{\tabcolsep}{4pt}\setlength{\LTcapwidth}{\linewidth}
\begin{longtable}{rL{.64\linewidth}l}
\caption{Original upstream Lean environments for every current project, ordered as in the publication concordance. Unmarked entries use an import pin or explicit original import report. A dagger marks the latest upstream default-branch revision before registration, used when the import pin is unavailable. Unpinned means the retained source or original import report records no upstream toolchain pin. Links lead to the source evidence.}\label{tab:source-versions}\\
\toprule No. & Project & Upstream Lean\\\midrule\endfirsthead
\toprule No. & Project & Upstream Lean\\\midrule\endhead
\bottomrule\endfoot
1 & 2-Coloring Cycles in One Round & \href{https://raw.githubusercontent.com/suomela/2-coloring-1-round/aa04b54db9a48d7a5084e263c147607af7744498/lean-toolchain}{4.28.0$^{\dagger}$}\\\addlinespace
2 & A 4AP-free permutation of the positive integers & \href{https://raw.githubusercontent.com/boonsuan/4ap/4630aefbbcb7f6537aceaeb1afac8dd3af455871/lean-toolchain}{4.33.1}\\\addlinespace
3 & A complex structure on the six-sphere & \href{https://github.com/plby/HopfProblem/blob/9ac8a456b526527837d7082ff775213ca8bc9809/lean-toolchain}{4.33.0}\\\addlinespace
4 & A Conditional Fourteen-Point Case of Erd{\H{o}}s Problem 132 & \href{https://raw.githubusercontent.com/lyfar/erdos-132-moment-obstruction-lean/cefcada06dcf71143a9ace9386ba84610f114122/lean-toolchain}{4.32.0-rc1$^{\dagger}$}\\\addlinespace
5 & A Conditional Sieve Criterion for Twin Primes via Krafft Geometry & \href{https://raw.githubusercontent.com/ElNando888/KrafftSieve/d6a3b6c862662cd91f6e7faeee39e1057efed575/lean-toolchain}{4.32.0-rc1$^{\dagger}$}\\\addlinespace
6 & A Detailed Proof of the Chudnovsky Formula & \href{https://raw.githubusercontent.com/ldct/lean-eval-chudnovsky/4a7505195632d378c8b5cb888864b344621e1d4d/lean-toolchain}{4.32.0-rc1$^{\dagger}$}\\\addlinespace
7 & A formalization of Borel determinacy in Lean & \href{https://raw.githubusercontent.com/sven-manthe/A-formalization-of-Borel-determinacy-in-Lean/42bc874b2357ca7e7573b31854a0d09761e11e41/lean-toolchain}{4.28.0-rc1$^{\dagger}$}\\\addlinespace
8 & A sharp 5/8 bound for Erd{\H{o}}s Problem 865 & \href{https://github.com/Vilin97/lean-pool/pull/322}{4.28.0}\\
\end{longtable}\endgroup

\clearpage
\section{Imported projects and their publications}\label{app:project-papers}
\begingroup\small\setlength{\tabcolsep}{4pt}\setlength{\LTcapwidth}{\linewidth}
\begin{longtable}{rL{.25\linewidth}L{.64\linewidth}}
\caption{Complete concordance of the current archive and its publications. Each registered project has a row and a link to its upstream repository. Companion papers and manuscripts describe the formalized results or development; mathematical sources provide underlying results. Source records distinguish papers, books, manuscripts, and repository documentation. Projects removed before the observation are excluded.}\label{tab:project-papers}\\
\toprule No. & Imported project / upstream & Publications and mathematical sources\\\midrule\endfirsthead
\toprule No. & Imported project / upstream & Publications and mathematical sources\\\midrule\endhead
\bottomrule\endfoot
1 & \href{https://github.com/suomela/2-coloring-1-round}{2-Coloring Cycles in One Round} & Companion paper: 2-Coloring Cycles in One Round~\citep{poolpubf55199bfe124}.\\\addlinespace
2 & \href{https://github.com/boonsuan/4ap}{A 4AP-free permutation of the positive integers} & Companion paper: A 4AP-free permutation of the positive integers~\citep{poolupdate_four_ap_free_permutation}.\\\addlinespace
3 & \href{https://github.com/plby/HopfProblem}{A complex structure on the six-sphere} & Mathematical manuscript: A compact complex threefold fibred by tori over the projective line, and the six-sphere~\citep{alpoge2026sixsphere}.\\\addlinespace
4 & \href{https://github.com/lyfar/erdos-132-moment-obstruction-lean}{A Conditional Fourteen-Point Case of Erd{\H{o}}s Problem 132} & Mathematical source: Constructions of Maximum Few-Distance Sets in Euclidean Spaces~\citep{poolpub22722e224e9f}.\\\addlinespace
5 & \href{https://github.com/ElNando888/KrafftSieve}{A Conditional Sieve Criterion for Twin Primes via Krafft Geometry} & Companion preprint: Variational Optimisation of Spectral Sieve Quotients for Primes in Bounded Symmetric Intervals~\citep{poolpubfe0cdfaaf0d4}.\\\addlinespace
6 & \href{https://github.com/ldct/lean-eval-chudnovsky}{A Detailed Proof of the Chudnovsky Formula} & Mathematical source: A detailed proof of the Chudnovsky formula with means of basic complex analysis -- Ein ausf\"uhrlicher Beweis der Chudnovsky-Formel mit elementarer Funktionentheorie~\citep{poolpub15d349129463}.\\\addlinespace
7 & \href{https://github.com/sven-manthe/A-formalization-of-Borel-determinacy-in-Lean}{A formalization of Borel determinacy in Lean} & Companion paper: A formalization of Borel determinacy in Lean~\citep{manthe2026borel}.\\\addlinespace
8 & \href{https://github.com/mrricky22/erdos-865-lean}{A sharp 5/8 bound for Erd{\H{o}}s Problem 865} & No publication identified; see upstream documentation.\\
\end{longtable}\endgroup

\end{document}